\documentclass[10pt, a4paper]{article}
\pdfoutput=1
\usepackage{lrec2022} 
\usepackage{multibib}
\newcites{languageresource}{Language Resources}
\usepackage{graphicx}
\usepackage{soul}
\usepackage{times}

\usepackage{lipsum}
\usepackage{latexsym}
\usepackage{tabularx}
\usepackage{booktabs}
\usepackage{multirow}
\usepackage{algorithm}
\usepackage{algpseudocode}
\usepackage{float}
\usepackage{amsmath}
\usepackage{mathtools}
\usepackage{makecell}

\usepackage{hyperref}
\hypersetup{colorlinks,allcolors=black}

\usepackage{titlesec}
\titleformat{\section}{\normalfont\large\bfseries\center}{\thesection.}{1em}{}
\titleformat{\subsection}{\normalfont\SmallTitleFont\bfseries\raggedright}{\thesubsection.}{1em}{}
\titleformat{\subsubsection}{\normalfont\normalsize\bfseries\raggedright}{\thesubsubsection.}{1em}{}
\renewcommand\thesection{\arabic{section}}
\renewcommand\thesubsection{\thesection.\arabic{subsection}}
\renewcommand\thesubsubsection{\thesubsection.\arabic{subsubsection}}

\usepackage{epstopdf}
\usepackage[utf8]{inputenc}

\usepackage{hyperref}
\usepackage{xstring}

\usepackage{color}

\title{LARD: Large-scale Artificial Disfluency Generation}

\name{\parbox{0.8\textwidth}{\centering Tatiana Passali$^{\ast\dagger}$, Thanassis Mavropoulos$^{\ast}$, Grigorios Tsoumakas$^{\dagger}$, Georgios Meditskos$^{\ast\dagger}$, Stefanos Vrochidis$^{\ast}$}}

\address{\\$^{\ast}$Centre for Research and Technology Hellas, Thessaloniki, Greece \\ 
         {\{tpassali, mavrathan, gmeditsk, stefanos\}}@iti.gr \\ \\
         $^{\dagger}$School of Informatics, Aristotle University of Thessaloniki, Thessaloniki, Greece  \\ 
         {\{scpassali, greg, gmeditsk\}}@csd.auth.gr \\ }

\abstract{
Disfluency detection is a critical task in real-time dialogue systems. However, despite its importance, it remains a relatively unexplored field, mainly due to the lack of appropriate datasets. At the same time, existing datasets suffer from various issues, including class imbalance issues, which can significantly affect the performance of the model on rare classes, as it is demonstrated in this paper. To this end, we propose LARD, a method for generating complex and realistic artificial disfluencies with little effort. The proposed method can handle three of the most common types of disfluencies:  repetitions, replacements, and restarts. In addition, we release a new large-scale dataset with disfluencies that can be used on four different tasks: disfluency detection, classification, extraction, and correction. Experimental results on the LARD dataset demonstrate that the data produced by the proposed method can be effectively used for detecting and removing disfluencies, while also addressing limitations of existing datasets.
 \\ \newline \Keywords{disfluency detection, synthetic dataset, data augmentation}}

\begin{document}

\maketitleabstract

\section{Introduction}
\label{sec:introduction}
Virtual assistants and spoken dialogue systems are increasingly used in many applications, such as automated customer service~\cite{gnewuch2017towards,sheehan2020customer}, education~\cite{clarizia2018chatbot,heryandi2020developing} and healthcare~\cite{tian2019smart,sezgin2020readiness}. Recent systems, which are powered by powerful Deep Learning (DL) models~\cite{golovanov2019large,wu2019proactive,majumder2020interview,zhong2020towards,zhao2020knowledge}, led to significantly more accurate systems compared to previous approaches. However, despite the considerable progress in this area, many difficult problems still remain unresolved. One of them is the fluidity of spoken language, which often includes various errors and self-corrections made on the fly, rendering the deployment of such systems in the wild especially challenging.

Indeed, when humans recognize that they have made an error in their speech, they automatically attempt to correct it by editing, reformulating, or completely restarting themselves. This is part of the natural and intuitive process that happens subconsciously during a spontaneous human dialogue flow. These interruptions and self-corrections of a speaker's utterances are called \textit{disfluencies}~\cite{shriberg1994preliminaries} and are a common trait of spontaneous speech. There is a large number of studies about the structure of disfluencies and their influence on the conversation between humans in theoretical level ~\cite{sparks1994structure,shriberg1994preliminaries,plejert2004fix,colman2011distribution,emrani2019conversation} which led to the development of various practical applications and models for disfluency detection~\cite{Zayats2016Disfluency,wang2016neural,Wang_Che_Liu_Qin_Liu_Wang_2020,rocholl2021disfluency}.

A practical virtual assistance system should be able to seamlessly handle these cases of disfluency and understand the actual intention of the user. This includes promptly detecting the disfluency, categorizing it, and then appropriately correcting the input to the system. However, despite its importance in these situations, disfluency detection remains a relatively unexplored field. This can be attributed to a large degree to the lack of appropriate datasets for training disfluency detectors. It is worth mentioning that Switchboard~\cite{Godfrey1992Switch}, the largest dataset in the literature that includes - as a side task - disfluency detection, contains less than 40,000 examples of disfluencies, with most of them (more than 50\%)~\newcite{shriberg1996disfluencies} belonging to the most trivial class (simple repetitions)~\cite{Zayats2016Disfluency,Zayats2019Switch}. On the other hand, other datasets, such as Fisher~\cite{cieri2004fisher}, may contain a larger amount of data, which however are not annotated. As a result, they cannot be easily used to develop a system that would meet the aforementioned requirements. 

Therefore, the need for a dataset for fine-grained disfluency detection becomes obvious. However, manually collecting and annotating such datasets can be expensive, since it requires appropriately trained personnel. An option to overcome this limitation is to use theoretical knowledge regarding the structure of disfluencies to generate synthetic ones. However, this requires very carefully designed methods, since in many cases the difficulty and the variety of the synthetic examples can be low, as highlighted in~\newcite{gupta2021disfl}, which can negatively impact the performance of DL models.

Revisiting disfluency detection in the wild, we demonstrate that models trained on existing datasets fail to recognize simple types of disfluency. To overcome this limitation, we propose an automated method for generating a vast amount of synthetic disfluencies on English text based on existing textual corpora that contain only fluent text. More specifically, we start with a large-scale dialogue dataset that contains real dialogues. Then, we use advanced Natural Language Processing (NLP) tools that, along with carefully designed generation methods, ensure the production of realistic and coherent examples with high variety. The proposed method can handle the three most common types of disfluency (repetitions, replacements, and restarts)~\cite{chalak2015replacement,emrani2019conversation}, while it can also support a variety of different tasks. The code is publicly available\footnote{\href{https://github.com/tatianapassali/artificial-disfluency-generation}{https://github.com/tatianapassali/artificial-disfluency-generation}}.

The contribution of our work is summarised in the following:
\begin{itemize}
    \item  We propose a method for generating realistic and coherent disfluency datasets.
    \item  We release a new dataset for disfluency detection based on the proposed method.
    \item We experimentally evaluate our framework in synthetic and real-world datasets using state-of-the-art NLP tools.
\end{itemize}

The rest of the paper is organized as follows. Disfluencies in human dialogue are discussed in Section~\ref{sec:background}. Related work is reviewed in Section~\ref{sec:related_work}. The proposed method is introduced in Section~\ref{sec:dataset_generation}. Experimental results are provided in Section~\ref{sec:experimental}. Finally, conclusions and future research directions are discussed in Section~\ref{sec:conclusion}.

\section{Disfluencies in Human Dialogue}
\label{sec:background}
This section introduces the structure, as well as the different categories of disfluencies.
\subsection{Structure of Disfluencies}
\newcite{shriberg1994preliminaries} introduced a standard annotation scheme, called \textit{reparandum/interregnum}, for identifying disfluencies. This annotation scheme involves the following fragments: a) the \textit{reparandum}, b) the \textit{interruption point}, c) the \textit{repair}, and optionally d) the \textit{interregnum}, which is located right before the repair. 

The reparandum indicates the disfluent part of the utterance: the part that is not correct and must be replaced or ignored. Typically, the speaker attempts to rephrase, edit or restate the reparandum. The reparandum is usually short and involves 2 to 3 words. Even though the reparandum is usually considered as a ``rough'' copy of the repair, it can also be completely irrelevant with the repair.

The interruption point indicates the start of the repair, if no interregnum exists. If an interregnum exists, then the interruption point indicates the start of the interregnum, after which the repair follows. The interruption point does not indicate an actual word, it rather marks the moment of speech when a speaker realizes the error and initiates the correction process. In other words, the interruption point typically occurs right after the last word that is being told before the interruption of speech. 

The interregnum consists of repair cues, such as those shown in Table~\ref{tab:repair_cues_types}, 
that are typically used to fill the gap between the disfluent speech and the associated repair. These types of fillers are generally ignored from dialogue systems as they do not contain any useful information. Most of the time, the presence of a repair cue indicates the presence of a disfluency in the utterance. 
Interregnums are relatively easy to be detected as they are usually fixed phrases, in contrast to reparandums that require a deeper understanding of the complex dialogue flow. For this reason, many methods ignore the interregnum and focus only on the detection of the reparandum and the repair~\cite{Zayats2016Disfluency,Zayats2019Switch,jamshid2018disfluency,Bach2019NoisyBM}.

\begin{table}[hbt]
\caption{Different types of repair cues as described by\protect\newcite{shriberg1994preliminaries}}
\label{tab:repair_cues_types}
\centering
\begin{tabularx}{\columnwidth}{lX}
\toprule
\textbf{Repair cues} & \textbf{Examples} \\ \midrule\midrule
filled pauses & um, uh   \\ \midrule
editing phrases    & oops, no, sorry, wait, I meant to say \\  \midrule
discourse markers &  well, actually, okay, you know, I mean \\
\bottomrule
\end{tabularx}
\end{table}

We typically meet a repair in a speaker’s utterance either after the presence of an interregnum or directly after the interruption point. The repair is the corrected fragment of speech and is usually different from the reparandum. However, in some cases the repair can be exactly the same as the reparandum or even completely empty.

\begin{figure}
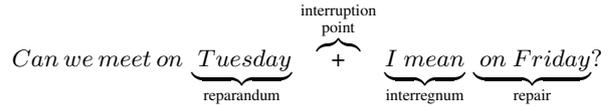


    \centering
    \small
    \[    Can \ we \ meet \ on \ \underbrace{ \ Tuesday}_{%
   \mathclap{\text{reparandum}}}   \ \overbrace{\text{+}}^{\substack{\text{interruption} \\ \text{point}}} \  \underbrace{\vphantom{\ Tuesday}I \ mean}_{%
   \mathclap{\text{interregnum}}} \ \underbrace{\ on \ Friday}_{%
   \mathclap{\text{repair}}} ? \] 
    \caption{An example of a disfluency, where the reparandum, interruption point, interregnum and repair are illustrated.}
    \label{fig:disfluency_example}
\end{figure}

An example of the aforementioned annotation scheme is described in Fig.~\ref{fig:disfluency_example}. In this example, the speaker mistakenly says Tuesday instead of Friday. The disfluent part of this example consists of the words \textit{``on Tuesday''} (reparandum), which is followed by an interregnum ({\em ``I mean''}) right after the interruption point and eventually by the correct choice of day (\textit{``on Friday''}), that makes the particular repair. To simplify the presentation of disfluencies under this scheme, the following notation is typically used to indicate the different parts of disfluencies:
\[ \text{[reparandum + \{interregnum\} repair] }\]
In this notation, the brackets (``['' and ``]'') are used to indicate the disfluency along with the repair.  The interruption point is indicated by the plus (``+'') symbol, while the interregnum, if present, is indicated by curly brackets (``\{'', ``\}'').

\subsection{Categories of Disfluencies}
\label{sec:categories}
Disfluencies can be categorized into three distinct classes based on~\newcite{shriberg1994preliminaries} speech repair typology: a) repetitions, b) replacements, and c) restarts, as described below:
\begin{itemize}
\item \textbf{Repetitions}: The speaker repeats a word, a phrase or a sequence of words. In repetitions, the reparandum and the corresponding repair are actually the same. This type of disfluency is the most common and is particularly easy to be detected~\cite{Zayats2014MultidomainDA,jamshid2019neural,Zayats2019Switch}. 
\item \textbf{Replacements}: The speaker replaces the disfluent word(s) or phrase with the fluent one. In replacements, the reparandum is replaced by the repair.
\item \textbf{Restarts}: The speaker abandons completely the initial utterance and restarts it. This type of disfluency does not actually involve a repair as the speaker begins a completely new utterance. 
\end{itemize}

Some examples of the different categories of disfluencies are shown in Table~\ref{tab:disfluency_types}.

\begin{table}[hbt]
\caption{Different types of disfluencies annotated based on the reparandum/interregnum scheme.}
\label{tab:disfluency_types}
\centering
\resizebox{\columnwidth}{!}{%
\begin{tabular}{ll}
\toprule
\textbf{Type}       & \textbf{Example}  \\ \midrule
repetition & Let's meet [today + today].     \\
replacement    & I want [the blue + \{no\} the red ] one.     \\
restart    & [Why don't you + ] I will do it later. \\
\bottomrule
\end{tabular}%
}
\end{table}

\section{Related Work}
\label{sec:related_work}
The related work is divided into three broad categories: a) models and methodologies for detecting and classifying disfluencies, b) existing datasets that can be used for training the aforementioned methods and c) data augmentation methods that attempt to alleviate the issues arising from the nature of existing datasets (e.g., limited size and variety) for tackling these tasks.

\subsection{Methodologies and Models}
Current research in disfluency detection is based on four methodologies: \textit{parsing-based}, \textit{noisy-channel}, \textit{sequence tagging} and \textit{disfluency correction} approaches. Parsing-based approaches work by identifying both the syntactic structure and the disfluencies in a text sequence~\cite{honnibal2014joint,rasooli2015yara,wu2015efficient,yoshikawa2016joint,jamshid2019neural}. Noisy-channel methods~\cite{charniak2001edit,johnson2004noisy,zwarts2010detecting} compute the similarity between the reparandum and the repair of an utterance in order to indicate a disfluency. \newcite{jamshid2017disfluency} combine a noisy-channel model to extract the best candidates for disluency analyses while a Long Short-Term Memory (LSTM) model is used to score these candidates and conclude to the most plausible one.

Sequence tagging models detect disfluencies in dialogues by predicting a tag/label to classify tokens as fluent or disfluent. Sequence tagging methodologies include Conditional Random Fields (CRF)~\cite{Georgila2009,ostendorf2013sequential}, Hidden Markov Models (HMMs) ~\cite{Liu2006enriching}, Semi-Markov CRF~\cite{ferguson2015disfluency}, Auto-Correlation CNNs~\cite{jamshid2018disfluency}, RNNs~\cite{Hough2015incremental,Zayats2016Disfluency,wang2016neural}, Transformers ~\cite{dong2019adapting,Wang_Che_Liu_Qin_Liu_Wang_2020} and on-device lightweight BERT architectures~\cite{rocholl2021disfluency}. 

Finally, a recent approach for disfluency correction, which was inspired by machine translation models, tackles disfluency detection as a sequence to sequence task where the disfluent sequence is ``translated'' into a clean and corrected version~\newcite{saini2021disfluency}.

\subsection{Datasets}
One of the most popular datasets that is used for training and evaluating the aforementioned models is Switchboard~\cite{Godfrey1992Switch}. Switchboard consists of 2,400 transcriptions of two-sided telephone conversations and approximately 190K utterances about a specific topic from a pre-defined list of 70 different topics. Switchboard is annotated with different types of speech tags and also contains metadata about some characteristics of the participants such as gender, education, area, etc. 
Disfluencies were annotated manually following the reparandum/interregnum structure ~\cite{shriberg1994preliminaries} that it will be extensively described in Section~\ref{sec:dataset_generation}.

Even though Switchboard is the most commonly used dataset for disfluency detection, only 5.9\% tokens of the dataset are actually annotated as disfluent ~\cite{charniak2001edit}. In addition, more than 50\% of the disfluency examples are classified as repetitions \cite{shriberg1996disfluencies}, which is considered as the easiest type of disfluency to identify \cite{Zayats2016Disfluency,Zayats2019Switch}.

Fisher English Training Transcripts \cite{cieri2004fisher} corpus is another dataset that contains disfluencies and consists of 5,850 telephone conversations and approximately 1.3M utterances between two speakers, compiled by Linguistic Data Consortium (LDC)\footnote{\href{https://www.ldc.upenn.edu/}{https://www.ldc.upenn.edu/}}. 
Each call, which lasts up to 10 minutes, is about an assigned topic, from a randomly generated list of topics. Some of the assigned topics are inherited from Switchboard corpus. Even though Fisher is clearly more extensive than Switchboard, it does not provide annotations for the existing disfluencies. 

Last but not least, \newcite{gupta2021disfl} recently released a new disfluency dataset for question answering of 12,000 
human-annotated disfluent questions.

\subsection{Data Augmentation}
The existing DL methods for disfluency detection and classification critically rely on the quantity and quality of the annotated training data, which can be a limiting factor as discussed. Some limited steps to alleviate this problem have been made~\cite{wang2018semi} using semi-supervised techniques leveraging knowledge from unlabeled datasets. \newcite{jamshid2019neural} use a self-training augmentation technique to obtain silver annotations for the Fisher~\cite{cieri2004fisher} dataset using a trained model on the Switchboard~\cite{Godfrey1992Switch} to augment the training data. \newcite{yang2020planning} use a simple planner-generator model in order to generate disfluent text, where a planner decides where to place the disfluent text which is created by the generator. A multi-task self-supervised learning method was proposed in~\newcite{Wang_Che_Liu_Qin_Liu_Wang_2020} where two self-supervised pre-training tasks were employed: a) a \textit{tagging} task and b) a \textit{sentence classification} task. At the same time, a pseudo-training dataset was created by randomly repeating, inserting or removing tokens from unlabeled sentences. The added tokens were randomly selected, thus the semantic consistency between the created disfluent sentence and the fluent one can not be guaranteed. Thus, unlike the proposed method, the aforementioned approaches either rely on very simple rules and techniques that are not capable of generating all different kinds of disfluencies that can be encountered in real scenarios. The proposed method is carefully designed to generate different types of disfluencies while preserving the semantic consistency between the fluent and the generated disfluent sentences, being closer to the disfluencies that can occur in a natural dialogue flow. 

\section{The LARD Method}
\label{sec:dataset_generation}
This section describes our Large-scale ARtificial Disfluency (LARD\footnote{We can poetically view disfluencies as the {\em lard} that on one hand, you know, spices up human dialogues, but which on the other hand should be removed to improve their vigor.}) method for generating disfluencies. In addition, it presents the large-scale dataset that we generated based on this method.

\subsection{Generation Pipeline}
LARD synthesizes disfluencies that belong to one of the three different types that were presented in Section \ref{sec:categories}: repetitions, replacements, and restarts. It is carefully designed to generate coherent and realistic synthetic examples, in order to ensure that the trained DL models will correctly generalize on real examples.

\begin{algorithm}
\caption{Repetition algorithm}\label{alg:repetitions}
\begin{algorithmic}[1]
\Require{(A fluent sequence $S_f$, the degree of repetition $d_r \in\{1,2,3\}$  )}
\Ensure{A disfluent sequence $S_{dis}$}
\Procedure{Create Repetitions}{$S_f$, $d_r$}
    \State $l_{s} \gets length(S_{f})$
    \State $random_{idx} \gets uniform(0, l_{s}-d_r$)
    \State $S_{dis} \gets$ Repeat the subsequence of tokens in $S_{f}$ starting from index $random_{idx}$ to degree $d_r$.
\State \Return $S_{dis}$
\EndProcedure
\end{algorithmic}
\end{algorithm}

The approach used for generating repetitions is shown in Algorithm~\ref{alg:repetitions}. More specifically, we select randomly an index from a sequence and we extend this sequence by repeating once the token in the corresponding index. Following this pattern, we create repetitions sub-cases for one, two and three consecutive words. For example, given a fluent sequence $S_f$ = ``I need to find a flight'', a possible disfluent sequence that includes one-word repetition will be $S_{dis}$ = ``I need to [find + find] a flight'' where the word ``find'' in position 4 is repeated. Similarly, given the same sequence for generating a two-word repetition, a possible generated disfluent sequence will be $S'_{dis}$ = ``[I need + I need] to find a flight'', where the two consecutive words ``I'' and ``need'' are repeated.

Similarly, the approach used for generating replacements is shown in Algorithm~\ref{alg:replacements}. In particular, we extract all the nouns, verbs and adjectives with NLTK ~\cite{bird2009natural} POS tagger, for each sequence. We also create a list that contains different repair cues including all of the fixed words of Table~\ref{tab:repair_cues_types}, apart from filled pauses. We add some variations of these words and phrases such as ``no, wait'', ``I am sorry'', ``no I meant to say'', ``no wait a minute'', ``well I actually mean'' etc. We do not include filled pauses in this list because these types of repair cues can easily be identified and removed from recent dialogue generation models. 

\begin{algorithm}[hbt]
\caption{Replacement algorithm}\label{alg:replacements}
\begin{algorithmic}[1]
\Require{(A fluent sequence $S_f$, part-of-speech $s_{pos} \in$ \{noun, verb, adjective\}, a boolean variable $cue$}
\Ensure{A disfluent sequence $S_{dis}$}
\Procedure{Create Replacement}{$S_f$, $cue$}
\If{there is $s_{pos}$ in $S_f$}
\State $W_{pos} \gets $ All the available words of the sequence $S_f$ that belong to the selected input part-of-speech $s_{pos}$
\State (a) Select randomly one of the available words from $W_{pos}$ as a repair candidate.
\State (b) Find all possible synonyms and antonyms for the selected repair candidate.
\State $S_{repl} \gets$ Select randomly an antonym or synonym as a reparandum candidate for the repair candidate
\State $idx_{s_{pos}} \gets$ index of the selected repair candidate in $S_f$
\State $d_{r} \gets uniform(0, idx_{s_{pos}})$ \Comment Define randomly the degree of replacement
\State (c) Repeat $d_{r}$ tokens before the $idx_{s_{pos}}$
\State (d) Place the reparandum candidate
\If{$cue$ == True} Randomly select a repair cue from list and place it after the reparandum candidate.
\EndIf
\State(e) Continue the rest of $S_f$.

\Else \State Select another sequence.
\EndIf
\EndProcedure
\end{algorithmic}
\end{algorithm}

We create simple one-word replacements or replacements that involve more than one word in both reparandum and repair. The steps for \textit{one-word replacement} generation are explained below:
\begin{enumerate}
    \item Given a sequence, we extract all the available words that belong to the selected part-of-speech for replacement, if exist.
    \item We randomly extract a repair candidate, a word whose POS is noun, verb or adjective, according to the desired POS.
    \item We generate a list of possible synonyms and antonyms for the selected repair candidate word using WordNet~\cite{bird2009natural}.
    \item We randomly select a synonym or antonym from the list as a reparandum candidate. The reparandum candidate (synonym or antonym) is placed in the reparandum right before the repair candidate. Note that we do not place it after the repair candidate in order to ensure that the fluent version of the sequence will be realistic. The selected repair candidate involves always only one word, however the reparandum candidate can vary from 1 to 4 words.
    \item Optionally, we add randomly a word or phrase from the list of repair cues as interregnum depending on the sub-class that we generate (i.e., with or without repair cue).
    \item Finally, the repair of the reparandum is the initially selected repair candidate.
\end{enumerate}

For example, given a fluent sequence $S_f$ = ``Find me a different one'', a possible disfluent sequence that includes an one-word replacement - the simplest form of a replacement - will be $S_{dis}$ = ``Find me a [same + \{sorry\} different] one''. In this example, the selected repair candidate is the adjective ``different'' and the adjective ``same'' is generated as a possible antonym (reparandum candidate) for the selected repair candidate. Then, ``same'' is replaced by the adjective ``different'', which was initially in the fluent sequence. Note that between the replacement and the repair, we randomly place a repair cue (i.e., ``sorry'') from a fixed list with possible repair cues. 

For replacements that involve \textit{more than one word} in both reparandum and repair, we repeat the same steps, but we also include some previous tokens from the sequence before the reparandum candidate. To do that we randomly select a degree $d_r$  for the replacement in order to define the number of tokens that will be repeated in the reparandum. The same tokens are repeated in the repair right before the selected repair candidate. For example, given another sequence $S'_f$ = ``I 'm looking for a salon in San Mateo'' with degree $d_r = 1$ a possible generated disfluent sequence will be $S'_{dis}$ = ``I 'm looking for [a beauty shop + \{I mean\} a salon] in San Mateo''. This is a more complicated example of the replacement class that involves more than one word in the reparandum. In particular, the noun ``beauty shop'' is generated as a synonym of the selected repair candidate ``salon'' and it is placed in the reparandum along with the previous token "a" ($d_r = 1$). Then, the subsequence ``a beauty shop'' is replaced by the subsequence ``a salon'', after the phrase ``I mean'', that is placed in the interregnum.

Eventually, the replacement class includes six different sub-classes: a) noun replacement with repair cue, b) noun replacement without repair cue, c) verb replacement with repair cue, d) verb replacement without repair cue, e) adjective replacement with repair cue and f) adjective replacement with repair cue. 

Finally, Algorithm~\ref{alg:restarts} is used for the generation of restarts. This algorithm assumes the existence of a fluent set that consists of two or more fluent sequences. First, we randomly select two sequences from the corresponding fluent set. Then, we split the first one in a random position and we combine the broken sequence together with the second unbroken one. For example, given a fluent sequence $S_{f1}$ = ``Do you want to check out on March 11th?'' and a fluent sequence $S_{f2}$ = ``When is the check-out date?'', a possible disfluent sequence that includes a restart will be $S_{dis}$ = ``[Do you want to + ] When is the check-out date?''. In this example, the initial fluent sequence $S_{f1}$ is randomly broken in a position and the whole fluent sequence $S_{f2}$ is placed right after the interruption point. Here, the repair is empty, since the speaker abandons the initial sequence and begins a new one. During this process, we make sure that we do not select two same sequences and that the beginning of the second sequence is different from the first broken sequence. This avoids the accidental generation of repetition examples, instead of restarts. 

\begin{algorithm}[!t]
\caption{Restart algorithm}\label{alg:restarts}
\begin{algorithmic}[1]
\Require{Two fluent sequences $S_{f1}$ and $S_{f2}$}
\Ensure{A disfluent sequence $S_{dis}$}
\Procedure{Create Restarts}{$S_{f1}$, $S_{f2}$}
\If{$S_{f1} \neq S'_{f2}$}
    \State $S'_{f1} \gets$ $S_{f1}$ broken in a random position. 
    \If{$S'_{f1} \neq$ beginning of $S_{f2}$}
    \State $S_{dis} \gets join(S'_{f1}, S_{f2})$
\EndIf
\Else \State Pick another sequence $S_{f2}$
\EndIf
\State \Return $S_{dis}$
\EndProcedure
\end{algorithmic}
\end{algorithm}

\subsection{Dataset Generation}
To construct artificial disfluencies based on LARD, we use an existing large-scale dialogue dataset: the Schema-Guided Dialogue (SGD) dataset~\cite{rastogi2019towards}. SGD is a multi-domain task-oriented dataset that contains approximately 18K conversations between users and virtual agents with more than 300K total individual utterances. It covers a large number of different domains including conversations for various typical scenarios, such as hotel reservation, flight booking, finding a movie and weather prediction.

SGD is selected as a basis for our synthetic dataset for two main reasons. First, it contains a large number of available dialogues that can be used to generate synthetic disfluencies with a large variety. This is especially critical given the data-demanding nature of DL models that rely significantly on annotated data. Second, the existence of different multi-domain conversations in SGD allows for training more robust models that perform well on different tasks and settings. 

Dataset generation is summarized as follows. First, we extract all the conversations from the SGD training set and split each conversation into sequence-level. Then, the total number of examples is split into four equal parts. The first part of the dataset remains unaltered, in order to include fluent sequences in our synthetic dataset. The others three parts are used for generating repetitions, replacements and restarts, respectively.

After applying this process on the SGD dataset, a total of 95,992 disfluencies were generated. Table~\ref{tab:statistics} presents the statistics of the generated data. The LARD dataset is balanced among the four different possible classes (fluency, replacement, repetition and restart). We split the dataset into 57,595, 19,198 and 19,199 examples for training, validation and testing, respectively. The dataset is publicly available at Zenodo\footnote{\href{https://zenodo.org/record/6451984}{https://zenodo.org/record/6451984}
}.

\begin{table}[hbt]
\caption{Statistics of the LARD dataset.}
\label{tab:statistics}

\centering
\begin{tabularx}{\columnwidth}{ll}
\toprule
\textbf{Dataset Statistics} \\ \midrule
\# repetitions (one word) & 8035     \\
\# repetitions (two words)    & 7964    \\
\# repetitions (three words)    & 7999 \\ \midrule
\# \textbf{repetitions} & \textbf{23998} \\ \midrule
\# noun replacements & 5023 \\
\# noun replacements + repair cue & 4900 \\
\# verb replacements  & 4910 \\
\# verb replacement + repair cue & 4814 \\
\# adjective replacements & 2098 \\ 
\# adjective replacements + repair cue & 2253 \\ \midrule
\# \textbf{replacements} & \textbf{23398} \\ \midrule
\# \textbf{restarts} & \textbf{23998} \\ \midrule
\# \textbf{fluencies} & \textbf{23998} \\ \midrule \midrule
\# \textbf{total examples} & \textbf{95992} \\ \bottomrule
\end{tabularx}

\end{table}

We create two different types of annotations. At the sentence-level, we define four different classes: fluency, repetition, replacement and restart. We also mark each token as fluent (repair or any other word outside the reparandum and interregnum) or disfluent (any word inside reparandum and interregnum).

\section{Empirical Evaluation}
\label{sec:experimental}
We present evaluation results on the proposed dataset in four different disfluency tasks: a) disfluency detection, b) disfluency classification, c) disfluency extraction, and d) disfluency correction. For disfluency classification, we consider all four classes (fluency, repetition, replacement and restart), while for disfluency detection we merge the repetition, replacement and restart classes into a disfluency class. For disfluency extraction, we classify each token of the sequence as fluent or disfluent. For disfluency correction, we use the synthetic sequence as input and the original fluent sequence as the desired output. Finally, we examine the behavior of pre-trained models fine-tuned on existing state-of-the-art datasets for disfluency detection like Switchboard. We demonstrate that even though models fine-tuned on Switchboard are particularly successful in detecting disfluencies like repetitions, they actually fail to recognize more complex types such as restarts and replacements.

\subsection{Experimental Setup}
For all the experiments we use Transformer-based models from the Hugging Face library~\cite{wolf2020transformers}. For the first three tasks, we use the pre-trained BERT base uncased model~\cite{devlin2019bert}, which consists of 12 encoder layers with 12 attention heads. This model is trained on large lower-cased English corpora: English Wikipedia and BookCorpus~\cite{zhu2015aligning}. We fine-tune this model on the LARD dataset for each task by adding a binary, multi-class and sequence classification head, respectively. For disfluency correction, we use T5-base~\cite{RaffelExploring2020}, which is a sequence-to-sequence model for text generation, with 12 encoder and decoder layers and 12 attention heads trained on Colossal Clean Crawled Corpus (C4). 

We set the learning rate to 0.00005 and fine-tune the pre-trained models for 20,000 steps with 16 batch size. Note that we do not extensively fine-tune these models, as this lies beyond the scope of this paper. Further fine-tuning and more extensive hyper-parameter searching could potentially lead to fine-tuned models with improved accuracy.

\subsection{Experimental Results}
The results for all the models on the four different tasks fine-tuned on the LARD dataset are shown in Table~\ref{tab:synthetic_results}. For all the classification tasks, we report precision, recall and f-measure, while for disfluency correction, we report BLEU~\cite{papineni2002bleu}. We can see that after fine-tuning on existing pre-trained models, we can easily reach a high accuracy on the LARD test set detecting, identifying and removing successfully different types of disfluencies.   

\begin{table}[tb]
\caption{Experimental results on models fine-tuned on LARD dataset. We report precision (prec), recall (rec) and f-measure (f1) (\%) for the classification tasks and BLEU (\%) for the  correction task.}
\label{tab:synthetic_results}
\resizebox{\columnwidth}{!}{%
\begin{tabular}{rllll} 
\toprule
\textbf{LARD dataset} & \textbf{Prec}  & \textbf{Rec}   & \textbf{F1}   & \textbf{BLEU}  \\ \midrule
Detection      & 97.63 & 97.61 & 97.62 & - \\
Classification & 97.31 & 97.30  & 97.29 & - \\
Extraction       & 98.12 & 96.6  & 97.30  &  - \\
Correction         & -     & -     & -    & 86.48 \\
\bottomrule
\end{tabular}}
\end{table}

\begin{table}[tb]
\caption{Experimental results of models fine-tuned on Switchboard.}

\label{tab:switchboard_results}
\resizebox{\columnwidth}{!}{%
\begin{tabular}{rllll} 
\toprule
\textbf{Switchboard}   & \textbf{Prec}  & \textbf{Rec}   & \textbf{F1}   & \textbf{BLEU}  \\ 
\midrule
Detection      & 94.44 & 94.31 & 94.37 & - \\
Extraction & 90.02 & 86.91  & 88.44 & - \\
Correction         & -     & -     & -    & 71.49 \\
\bottomrule
\end{tabular}}
\end{table}

We also examine the behavior of models with the same architecture fine-tuned on the Switchboard dataset~\cite{Godfrey1992Switch,Zayats2019Switch}, following the common train/dev/test splits, as they are established from~\newcite{charniak2001edit}. The experimental results reported in Tables~\ref{tab:synthetic_results} and~\ref{tab:switchboard_results} reveal that models trained on Switchboard perform worse (on the corresponding test set) compared to models trained on LARD. This observation led us to further examine the behavior of models fine-tuned on Switchboard using the LARD test set. 

The experimental results are reported in Table~\ref{tab:switchboard_analysis}, where the accuracy for different disfluency classes is reported. Even though models fine-tuned on Switchboard can achieve high accuracy in the repetition class, their performance significantly degenerates in the replacement ({$\sim$}55\%) and restart classes ({$\sim$}20\%).
On the contrary, the model fine-tuned with the proposed method on the detection task achieves over 95\% accuracy on all classes, while the one fine-tuned on the more challenging classification task, achieves over 93\% accuracy on all classes. This finding can be attributed to the large number of repetitions included in Switchboard~\cite{shriberg1996disfluencies}, along with the significantly smaller number of replacements and restarts. This finding also demonstrates the potential of the LARD method for generating complex disfluencies, allowing for increasing models' performance and robustness compared to existing and more imbalanced datasets.




\begin{table}[tb]
\caption{Accuracy (\%) for different disfluency classes (repetitions, replacements and restarts) and models trained on different datasets. 
Switchboard dataset supports only detection task. LARD method supports both detection and classification tasks.}

\label{tab:switchboard_analysis}
\resizebox{\columnwidth}{!}{%
\begin{tabular}{rll|l} 
\toprule
               & \textbf{Switchboard}  & \textbf{LARD} & \textbf{LARD}   \\ 
               & \textbf{(detection)}  & \textbf{(detection)} & \textbf{(classification)}   \\ 
               \midrule
Repetitions  & 85.42 & 99.57  & 99.50  \\
Replacements & 54.52 & 99.67  & 98.39 \\
Restarts     & 19.6  & 95.08  & 93.89 \\
\bottomrule
\end{tabular}}
\end{table}
\section{Conclusion}
\label{sec:conclusion}
We proposed LARD, a carefully designed mechanism for generating large-scale synthetic disfluencies from fluent text. We also compiled a large-scale dataset based on the proposed method which consists of approximately 96K examples, building upon an existing dialogue dataset: SGD~\cite{rastogi2019towards}. The generation pipeline is carefully designed in order to create complex, yet realistic and coherent disfluencies from fluent sequences. Experiments on the generated synthetic dataset using the LARD method on four different tasks (disfluency detection, classification, extraction and correction) showed that fine-tuned models can be effectively used to detect, identify and remove disfluencies from disfluent sequences. Furthermore, the thorough analysis conducted on models fine-tuned both on the generated dataset and Switchboard dataset revealed a significant weakness of the latter, arising from the imbalanced number of training examples among different classes, which can be successfully addressed using the proposed method.

Several interesting future work directions exist, including extending the proposed method to generate multiple disfluencies for training models that can handle these more difficult cases. Also, the proposed method highlights another research direction for modeling other linguistic phenomena, such as coreference resolution~\cite{lee2018higher} and discourse relations~\cite{qin2017joint}, e.g., contrast, elaboration, clarification, and others, providing an efficient way for generating highly realistic datasets that include them, based on existing large-scale fluent datasets.

\section{Acknowledgements}
This work is partially funded by the European Commission as part of its H2020 Programme, under the contract number 870930-IA (WELCOME).




\section{Bibliographical References}\label{reference}

\bibliographystyle{lrec2022-bib}
\bibliography{lrec2022-example}


\end{document}